\documentclass[letterpaper]{article} 
\usepackage{aaai2026}  
\usepackage{times}  
\usepackage{helvet}  
\usepackage{courier}  
\usepackage[hyphens]{url}  
\usepackage{graphicx} 
\usepackage{natbib}  
\usepackage{caption} 
\usepackage{algorithm}
\usepackage{amsmath}
\usepackage{amssymb}
\usepackage{booktabs}
\usepackage{arydshln}
\usepackage{algorithm}
\usepackage{algpseudocode}

\usepackage{newfloat}
\usepackage{listings}
\DeclareCaptionStyle{ruled}{labelfont=normalfont,labelsep=colon,strut=off} 
\floatstyle{ruled}
\newfloat{listing}{tb}{lst}{}
\floatname{listing}{Listing}
\title{DECOR: Deep Embedding Clustering with Orientation Robustness\thanks{This paper was accepted to the KGML Bridge at AAAI 2026 (non-archival)}}
\author {
    Fiona Victoria Stanley Jothiraj\textsuperscript{\rm 1}\thanks{Work conducted as part of 2025 summer internship at Micron Technology.},
    Arunaggiri Pandian Karunanidhi\textsuperscript{\rm 2},
    Seth A. Eichmeyer\textsuperscript{\rm 2}
}
\affiliations {
    \textsuperscript{\rm 1}Oregon State University\\
    \textsuperscript{\rm 2}Micron Technology\\
    stanleyf@oregonstate.edu, akarunanidhi@micron.com, saeichmeyer@micron.com
}

\begin{document}

\maketitle

\begin{abstract}
In semiconductor manufacturing, early detection of wafer defects is critical for product yield optimization. However, raw wafer data from  wafer quality tests are often complex, unlabeled, imbalanced and can contain multiple defects on a single wafer, making it crucial to design clustering methods that remain reliable under such imperfect data conditions. We introduce DECOR, a \textit{deep clustering with orientation robustness} framework that groups complex defect patterns from wafer maps into consistent clusters. We evaluate our method on the open source MixedWM38 dataset, demonstrating its ability to discover clusters without manual tuning. DECOR explicitly accounts for orientation variations in wafer maps, ensuring that spatially similar defects are consistently clustered regardless of its rotation or alignment. Experiments indicate that our  method outperforms existing clustering baseline methods, thus providing a reliable and scalable solution in automated visual inspection systems.
\end{abstract}


\section{Introduction}
Clustering is an unsupervised learning technique used to group unlabeled data based on the underlying similarity in their feature representations. Methods are often categorized into hard clustering and soft clustering based on the exclusivity of clusters: hard clustering assigns each data point to exactly one cluster, whereas soft clustering allows a data point to belong to multiple clusters with varying probabilities. When dealing with high-dimensional data such as wafer maps, autoencoder variants are commonly used to extract the low-dimensional latent embeddings that preserve structural information. These embeddings (often a 1-dimensional vector) is subsequently clustered using well-established approaches, including centroid-based clustering (Eg. K-Means), density-based clustering (Eg. DBSCAN, HDBSCAN, OPTICS), distribution-based clustering (Eg. Gaussian Mixture Models, Expectation-Maximization) or hierarchical clustering (Eg. Agglomerative Clustering). 

Despite their effectiveness and widespread use, most clustering methods rely on fixed assumptions, such as the number of clusters (k), maximum distance between points ($\epsilon$), and choice of distance metric (Eg. L1 or L2 distance). This reliance of assumptions makes it impractical for dynamic wafer inspection, where the wafer defect distribution can evolve over time, and clustering needs to be applied frequently to identify outlier or anomalous patterns without prior knowledge of their prevalence or scale.

Within semiconductor manufacturing, this challenge is compounded by the fact that wafer maps often exhibit orientation variability. The same defect pattern may appear at different rotations due to variations of wafer placement or handling. Standard clustering methods treat rotated instances as being distinct patterns and this leads to fragmented clusters. 

These challenges motivate the adoption of non-parametric methods of clustering \cite{nguyen2021clusformer, xing2021learning, mcconville2021n2d} in our work which can adaptively discover clusters in latent space without requiring prior specifications, while also aligning spatially similar wafer defects across orientations. We refer to our approach as DECOR (\textbf{D}eep \textbf{E}mbedding \textbf{C}lustering with \textbf{O}rientation \textbf{R}obustness).

\section{Method}
\begin{figure*}[h]
    \centering
    \includegraphics[width=\textwidth]{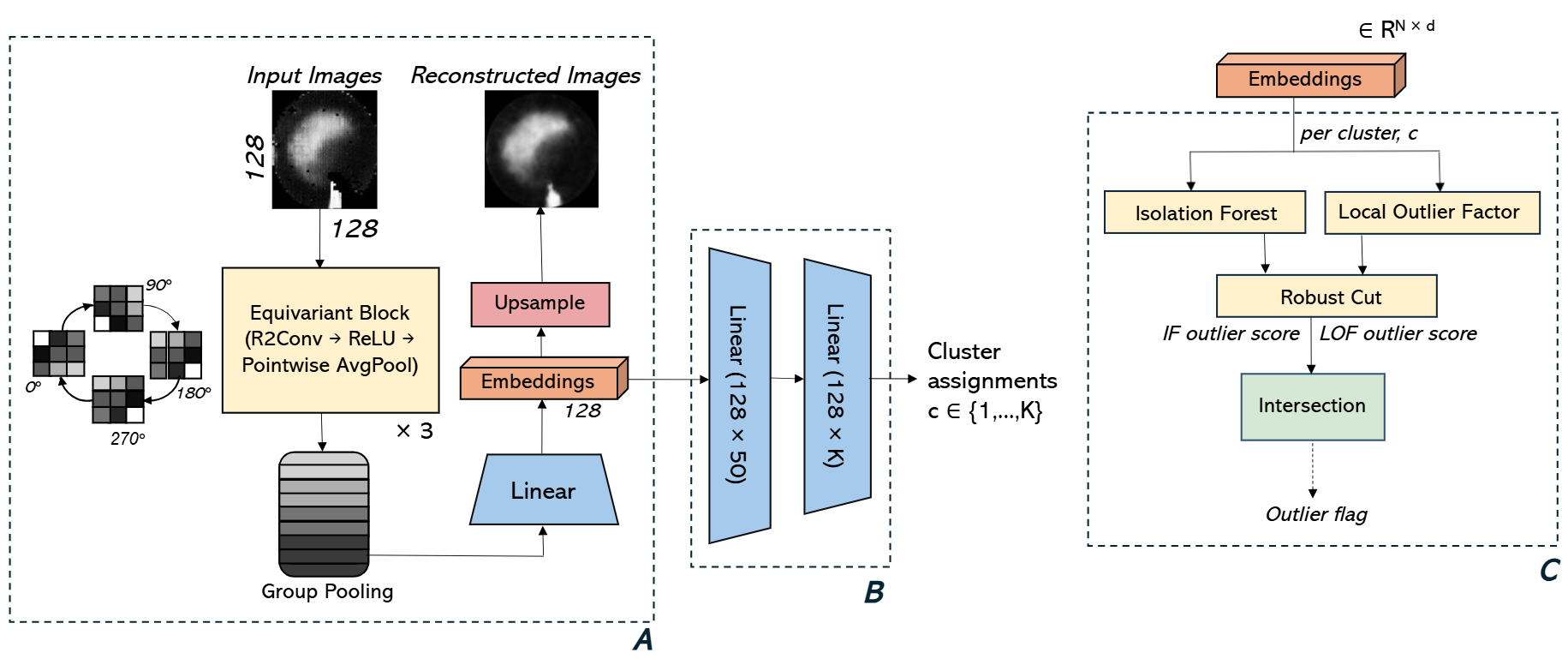}
    \caption{Overview of the workflow. Input wafer images are processed through an embedding extractor model, generating 128-dimensional embeddings along with their reconstructed images. These embeddings are then fed into a trained non-parametric clustering model based on DeepDPM \cite{Ronen:CVPR:2022:DeepDPM} to obtain cluster assignments. Finally, embeddings within each cluster are analyzed using an ensemble outlier detection algorithm to flag potential outliers.}
    \label{stage1-2-3}
    \end{figure*}

We propose a novel three-step approach (Algorithm \ref{alg:wafer_clustering}) for non-parametric wafer map clustering and cluster-aware outlier detection (Figure ~\ref{stage1-2-3}). Our approach consists of: (\textbf{A}) a rotation- and flip-invariant embedding extractor - RCAE, (\textbf{B}) a non-parametric clustering module, and (\textbf{C}) an ensemble outlier detection mechanism.

\begin{algorithm*}[t]
  \caption{DECOR: Dynamic Embedding Clustering with Orientation Robustness}
  \label{alg:wafer_clustering}
  \begin{algorithmic}[1]
    \Require Images $\mathcal{I} = \{I_1,\dots,I_N\}$; embedding dimension $d$; initial number of clusters $K_{\text{init}}$
    \Ensure Soft memberships $\mathbf{P}\in[0,1]^{N\times K}$; hard labels $\mathbf{c}\in\{1,\dots,K\}^N$; outlier set $\mathcal{O}^\ast$
    \For{$i=1$ to $N$}
      \State $I_i \gets \textsc{NormalizeAndMask}(I_i)$
      \State $z_i \gets \textsc{FeatureExtractor}(I_i)\in\mathbb{R}^{d}$
    \EndFor
    \State $\mathbf{Z} \gets [\,z_1;\dots;z_N\,]\in\mathbb{R}^{N\times d}$
    \State $\mathcal{A} \gets \textsc{DynamicClustering}(K_{\text{init}})$
    \State $\mathbf{P} \gets \mathcal{A}.\textsc{SoftCluster}(\mathbf{Z}) \in [0,1]^{N\times K}$
    \State $\mathbf{c} \gets \arg\max_{k\in\{1,\dots,K\}} \mathbf{P}$ \Comment{Hard labels from soft memberships}
    \State $\mathcal{O}^\ast \gets \bigcup_{k=1}^{K} \textsc{OutlierDetection}(\{z_i \mid c_i = k\})$ \Comment{Outlier detection per cluster}
    \State \Return $\mathbf{c},\, \mathcal{O}^\ast$
  \end{algorithmic}
\end{algorithm*}

    \subsection{Feature Extractor}
    To obtain robust and well-separated embeddings from high-dimensional wafer maps, we evaluate several architectures, including MoCo~\cite{he2019moco, chen2020mocov2, chen2020simple} for contrastive representation learning and convolutional autoencoders \cite{masci2011stacked} for reconstruction-based learning. While MoCo has shown effectiveness in general contrastive learning tasks, it lacks explicit reconstruction and its architecture does not inherently enforce symmetry handling. However, our experiments indicate that a rotation invariant convolutional autoencoder (CAE \cite{masci2011stacked}) with built-in symmetry handling produced more compact and separable clusters in the reduced embedding space (ie. RCAE). We processed wafer maps of size $128\times128$ through three \emph{rotation- and flip-equivariant} blocks, each comprising an \texttt{R2Conv} layer~\cite{e2cnn}, ReLU layer for non-linearity, and \texttt{PointwiseAvgPool} with stride~2. The blocks use increasing numbers of regular representation fields (8, 16, 32), achieving equivariance to the dihedral group $D_4$ - four discrete rotations (0$^\circ$, 90$^\circ$, 180$^\circ$, 270$^\circ$) and two mirror flips. To convert the equivariant features into orientation-invariant descriptors, we applied \texttt{GroupPooling}, which collapses all orientation channels within each field. A \texttt{Linear} layer then maps the resulting tensor to a 128-dimensional latent vector. The decoder mirrors the encoder, using three \texttt{ConvTranspose2D}+\texttt{ReLU} blocks and a final \texttt{Sigmoid} activation to reconstruct images within $[0,1]$.

    \subsection{Non-Parametric Clustering}
    For clustering in the latent space, we use DeepDPM \cite{Ronen:CVPR:2022:DeepDPM}, a non-parametric deep clustering framework based on the Dirichlet Process Mixture Model (DPMM). Unlike parametric clustering approaches like K-Means, which require a fixed number of clusters, DeepDPM infers the optimal number of clusters from the data distribution. The trained clustering model is implemented as a two-layer multilayer perceptron comprising a \texttt{Linear(128, 50)} layer followed by \texttt{Linear(50, $K$)}, where $K$ is adaptively determined during training. The model produces soft cluster membership probabilities, from which hard assignments are obtained.

    \subsection{Outlier Detection}
    In the latent space produced by the invariant encoder, outlier detection is performed with an ensemble of Isolation Forest (IF)~\cite{liu2008isolation, liu2012isolation} and Local Outlier Factor (LOF)~\cite{breunig2000lof}. IF provides a global, partition-based notion of isolation, while LOF captures local density deviations. Their raw scores are converted to outlier decisions using robust, per-cluster thresholding rule. 
    For the score vector \(s\), we define the decision threshold as
    \(\tau=\operatorname{median}(s)+k\cdot\operatorname{MAD}(s)\) where the Median Absolute Deviation (MAD) is given by \(\operatorname{MAD}(s)=\operatorname{median}(|s-\operatorname{median}(s)|)\). We use the median and MAD rather than mean and standard deviation to ensure robustness against heavy-tailed distributions and pre-existing outliers. To stabilize LOF across heterogeneous cluster sizes, the neighborhood size was chosen adaptively as \(k_{\text{LOF}}=\operatorname{clip}(\sqrt{N},\,k_{\min},\,k_{\max})\) for cluster cardinality \(N\). IF was trained with a conservative contamination prior (\texttt{hi\_cont} = 0.20) to avoid over-flagging, while the subsequent robust cut calibrates per-cluster selectivity. Final outlier labels were issued only when both detectors agree, \(\mathbf{1}_{\text{final}}=\mathbf{1}_{\text{IF}}\wedge\mathbf{1}_{\text{LOF}}\). This process empirically improves precision on rare, semantically novel wafer defects while retaining sensitivity through the complementary failure modes of the two detectors.
    
\section{Experiment}
    \subsection{Dataset}
    We train and evaluate our method on the opensource MixedWM38 \cite{wang2020deformable} dataset, which contains over 38{,}000 wafer maps with both single-defect and mixed-defect patterns, spanning more than 38 distinct defect pattern combinations. In particular, the dataset includes 1 normal pattern, 8 single-defect and 29 mixed-defect patterns. The base defect types are \emph{Center, Donut, Edge-Loc, Edge-Ring, Loc, Near-full, Scratch, and Random}. 
    To ensure consistent multi-label distribution across training and testing, we utilize MultilabelStratifiedShuffleSplit  \cite{sechidis2011stratification}, which preserves the proportion of each defect type in both subsets. Additionally, all wafer maps are resized to $128\times$128 pixels to balance reconstruction fidelity and clustering performance.

    \subsection{Results}
    We evaluate the clustering quality using standard metrics like: \textit{Normalized Mutual Information (NMI)} and \textit{Adjusted Rand Index (ARI)}. Since the ground truth (GT) is multi-label (i.e. single wafer can contain multiple defects), we perform a cluster-aware dominant label reduction of GT: for each wafer, if multiple defect labels are present, the most frequent within the sample's predicted cluster is assigned. NMI and ARI are then computed between these dominant labels and the predicted cluster assignment. For robustness, we repeat experiments with multiple random seeds and report mean values and standard error across three runs.

    A key challenge with this dataset is the absence of a well-defined ground-truth number of categories since wafers may contain overlapping or mixed defect types. This makes comparisons with parametric baselines (eg. $K$-Means) difficult, as their performance depends heavily on the choice of $K$.

\begin{figure*}[h]
        \centering
        \includegraphics[width=\textwidth]{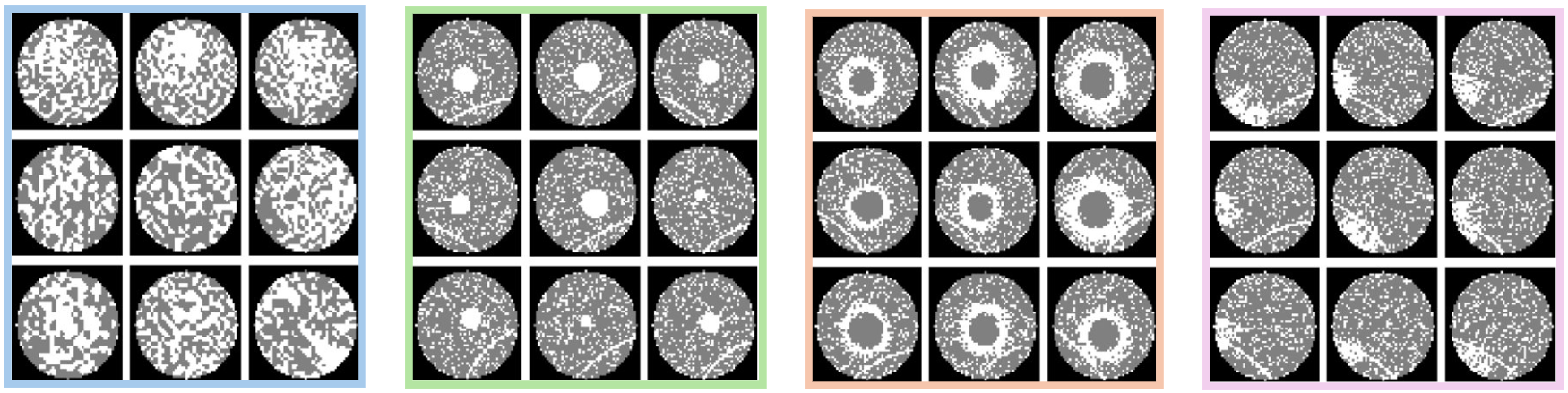}
        \caption{Examples of MixedWM38 images clustered together by DECOR. Each panel represents a different cluster. The grouping of images with varying rotational orientations (\textit{scratch, and local} pattern) indicates that DECOR exhibits rotational invariance.}
        \label{rcae_samples}
        \end{figure*}

    Table \ref{perf} reports the performance of different embedding-clustering combinations. Our results indicate that RCAE embeddings, when paired with a non-parametric clustering approach (DeepDPM), achieves highest performance in both NMI and ARI. We observe that RCAE+DeepDPM produces more compact and orientation-invariant clusters compared to baselines (Figure \ref{rcae_samples}). Furthermore, we find the quality of clustering correlates with the degree of separation between defect patterns in latent space: the more distinct the separation in 2D/3D projections (via \cite{mcinnes2018umap} \cite{maaten2008visualizing}), the better the clustering performance.
        \begin{table*}[h]
            \caption{Performance under different experiments indicating the mean$\pm \text{std. dev}$ across 3 runs. Methods marked with $^p$ are parametric (require \textit{K}). Experiment details are provided in \textit{Appendix A-C}}.
            \label{perf}
            \centering
            \begin{tabular}{llllll}
                \toprule
                \textbf{Embedding} & \textbf{Clustering} & \textbf{Final/Best \textit{K}} & \textbf{NMI $\uparrow$} & \textbf{ARI $\uparrow$}\\
                \midrule
                CAE & $K\text{-Means}^p$ & 24 & 0.503 ± 0.00 & 0.199 ± 0.00 \\
                MoCo & $K\text{-Means}^p$ & 18 & 0.409 ± 0.00 & 0.173 ± 0.01 \\
                RCAE & $K\text{-Means}^p$ & 30 & 0.529 ± 0.00 & 0.205 ± 0.00 \\
                \midrule
                CAE & DeepDPM & 25 & 0.498 ± 0.05 & 0.218 ± 0.01 \\
                MoCo & DeepDPM & 30 & 0.273 ± 0.00 & 0.117 ± 0.00 \\
                \textbf{RCAE (Ours)} & DeepDPM & 22 & 0.543 ± 0.03 & 0.296 ± 0.00 \\
                \bottomrule
            \end{tabular}
        \end{table*}
    
\section{Conclusion}
In this work, we address the challenges of reliable wafer map clustering under complex, noisy and multi-defect conditions. We propose DECOR, a novel framework that combines an orientation-invariant embedding extractor, a non-parametric clustering model and an outlier detection mechanism to group wafer maps into meaningful clusters, while flagging outlier defects. Experiments on the multi-label MixedWM38 dataset demonstrate that our method achieves superior clustering performance with minimal computational overhead, by leveraging lightweight yet effective models. One limitation of the non-parametric clustering approach is the necessity of multiple runs to determine an appropriate initial cluster count ($k_{\text{init}}$) and the optimal number of training epochs for DeepDPM clustering that is tailored to the dataset. Furthermore, evaluating clustering quality on multi-label datasets remains challenging, as metrics like NMI and ARI require simplification to dominant labels per image, which might not capture the full system’s performance. 

As part of future work, we plan to explore multi-label-aware clustering metrics, and integrate temporal modeling, leveraging soft cluster assignments produced by non-parametric clustering to track the evolution of defect patterns across production cycles. We also plan to benchmark DECOR against additional parametric clustering models, including Gaussian Mixture Models and deep clustering frameworks beyond K-means, to further assess its robustness and generalization. By advancing reliable unsupervised learning for imperfect and complex data, DECOR provides a foundation that generalizes beyond semiconductor wafers to other domains/datasets.

\appendix
\section{Experimental Setup}
We evaluate our method on the MixedWM38 wafer dataset using a multilabel stratified train-test split. After normalization, we applied Edge masking to prevent edge artifacts from dominating the embeddings. We also applied Gaussian blur transformation (kernel size 5, and sigma 1.0) to smooth noise and highlight the main defect patterns for clustering. Our embedding feature extractor was compared against baselines including vanilla CAE and MoCo, all trained under identical conditions for fair comparison. The RCAE model was trained using the Adam optimizer (initial learning rate: $1e^{-3}$), MSE loss, batch size of 128, and 1000 epochs. For DeeDPM clustering, we set hyper parameters as follows:  $\nu$ is d + 2 (where d is the input dimension to the clustering model or the latent dimension), $k_{init}$ is 30, and maximum epochs is 200. Although labels were withheld during training, we report evaluation metrics using NMI and ARI.

All experiments were implemented in PyTorch 2.6.0 and run on a single NVIDIA H100 GPU (80GB memory) with  32-core CPUs on a local compute cluster node.Training the RCAE model for 1000 epochs took approximately 6 hours per run, while training the non-parametric clustering model required about 2 hours. Across all experiments (including baselines), the total compute amounted to approximately 30GPU hours. Storage requirements were minimal, with about 400MB for datasets and about 1GB for checkpoints and logs.

    \begin{table*}[h]
        \caption{Implementation details of feature extractor models.}
        \label{implm}
        \centering
        \begin{tabular}{lll}
            \toprule
            \textbf{Architecture} & \textbf{Loss} & \textbf{lr} \\
            \midrule
            CAE: Conv(16) - Conv(32) - Conv(64) - Conv(256) - FC(128) & MSE & 0.001 \\
            MoCo: ResNet18 - FC(512) - FC(128) & InfoNCE & 0.03 \\
            RCAE: R2Conv(8) - R2Conv(16) - R2Conv(32) - GroupPool - FC(128) & MSE & 0.001 \\
            \bottomrule
        \end{tabular}
    \end{table*}

\section{Outlier Detection}
\subsection{Isolation Forest (IF) and Local Outlier Factor (LOF)}

Outlier detection is an essential task in many domains, especially in wafer map analysis where it is crucial to distinguish between genuine defects and noise. Isolation Forest (IF)~\cite{liu2008isolation, liu2012isolation} and Local Outlier Factor (LOF)~\cite{breunig2000lof} are both widely used for outlier detection due to their ability to handle high-dimensional and complex data. However, each has its strengths and limitations, which is why an ensemble approach is adopted to leverage the advantages of both.

Isolation Forest operates by randomly partitioning the feature space and isolating data points. Outliers, being different from normal points, are easier to isolate and therefore receive a lower score. On the other hand, Local Outlier Factor (LOF) focuses on local densities, comparing the density of a point with its neighbors. Points with significantly lower density than their neighbors are flagged as outliers.

While these algorithms are powerful on their own, they tend to excel in different types of data and outlier characteristics. To take advantage of both, an ensemble approach is employed, where outliers are flagged only if both Isolation Forest and LOF agree on the outlier, reducing the chance of false positives and increasing the robustness of the results.

\subsection{Robust Cut and Thresholding}

The performance of outlier detection algorithms depends heavily on defining appropriate thresholds for outlier scores. In this study, the Robust Cut approach is used to define these thresholds in a way that adapts to the data's underlying distribution. The Robust Cut uses the Median Absolute Deviation (MAD) to ensure that the threshold is not influenced by extreme values (outliers) in the score distribution.

The formula for the robust cut is defined as:
\[
\text{Threshold} = \text{Median}(scores) + k \times \text{MAD}(scores)
\]

where \(\text{Median}(scores)\) is the median of the outlier scores,
\(\text{MAD}(scores)\) is the median absolute deviation, calculated by \(\text{MAD} = \text{Median}(|\text{scores} - \text{Median}(scores)|)\),
and \(k\) is a scaling factor, which controls how aggressive the thresholding is.

    \begin{figure*}[h]
        \centering
        \includegraphics[width=0.6\textwidth]{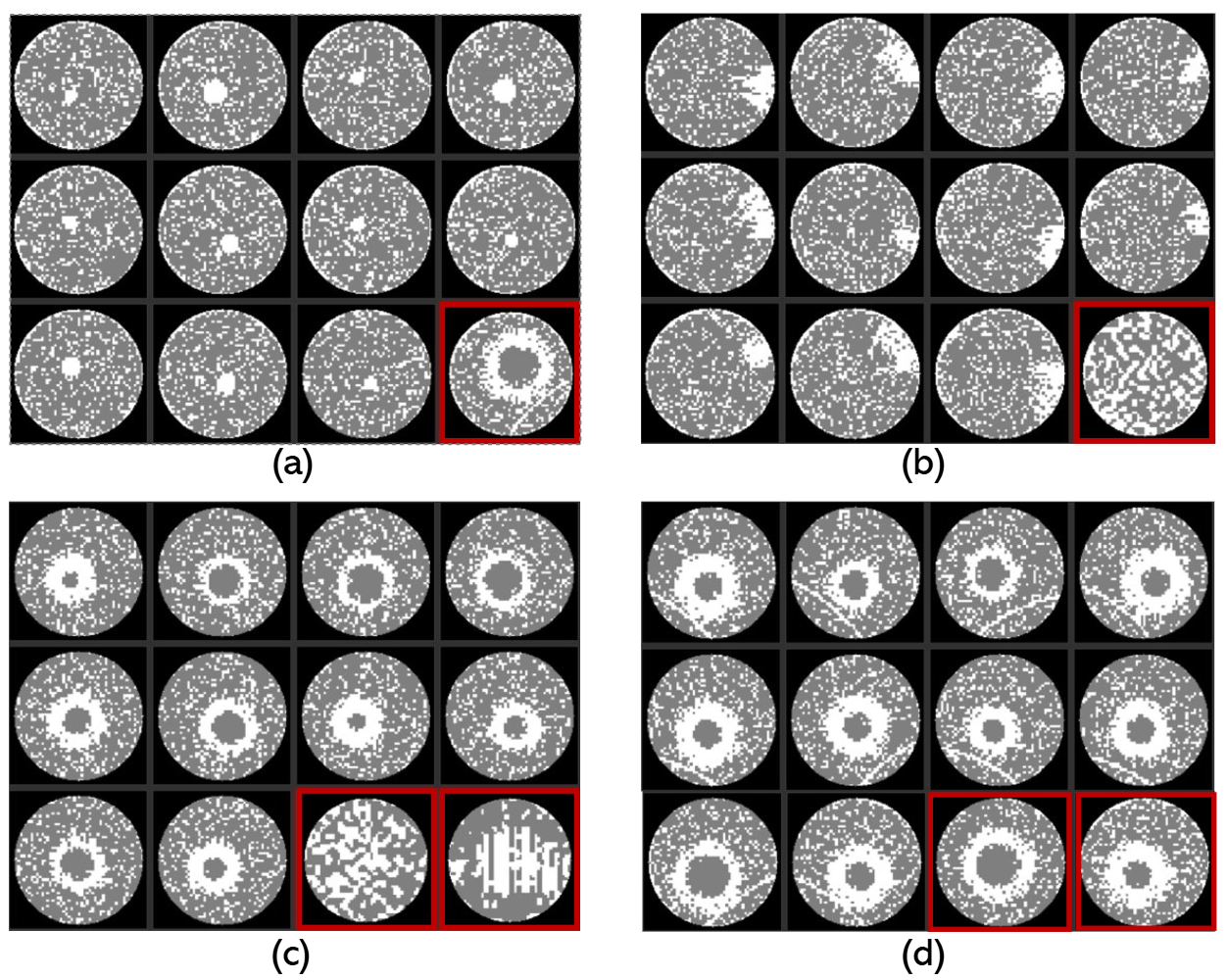}
        \caption{Examples of clusters produced by DECOR with detected outliers highlighted in red.
        (a) a \textit{center} defect cluster containing instances of \textit{donut} defect outliers
        (b) a \textit{local} defect cluster containing instances of \textit{random} defect outliers
        (c) a \textit{donut} defect cluster containing instances of \textit{random and nearfull} defect outliers
        (d) a \textit{donut + scratch} defect cluster containing instances of \textit{donut} defect outliers.}
        \label{outlier}
    \end{figure*}
    
Median is preferred over the mean because it is less sensitive to extreme values, which are common in high-dimensional data. This ensures that the threshold is robust to noise or outliers in the data and avoids flagging too many points as outliers when the dataset contains a few extreme cases.

MAD provides a more robust estimate of spread, which is particularly important when dealing with outliers. By scaling the MAD with \(k\), we can control how aggressively the algorithm flags outliers.

The contamination parameter (\(hi\_cont\)) is set to 0.20, meaning that 20\% of the data are expected to be outliers. This is a conservative choice to ensure that true outliers are not overlooked, especially in scenarios where outliers might be rare or subtle. Setting a low contamination value prevents the algorithm from being overly aggressive and helps in retaining the majority of the dataset as normal.

\subsection{Adaptive Thresholding and LOF Parameters}

The Adaptive k parameter in Local Outlier Factor (LOF) adjusts the number of neighbors considered when evaluating the density of each point. Instead of a fixed value for \(k\), we compute an adaptive \(k\) based on the size of the dataset - specifically the square root of the number of points in the cluster, constrained within a predefined range \([min\_k, max\_k]\). This adaptability ensures that the LOF algorithm remains sensitive to local structures regardless of cluster size, which is essential when dealing with clusters of varying densities.

The formula for the adaptive \(k\) is as follows:

\[
k = \text{Clip}(\sqrt{N}, min\_k, max\_k)
\]

where \(N\) is the number of data points in the cluster and \(\text{Clip}(x, min\_k, max\_k)\) ensures that \(k\) remains within the bounds of \([min\_k, max\_k]\).

By choosing the appropriate \(k\), LOF can accurately capture the local structure of the data and identify outliers that are dense in their neighborhoods but sparse relative to other areas.

\subsection{Ensemble Outlier Detection}

The ensemble approach combines the results of Isolation Forest (IF) and Local Outlier Factor (LOF) to determine whether an image is an outlier. The method is based on the assumption that true outliers are detectable both globally (via IF) and locally (via LOF), so an intersection of both methods is used for a more reliable result. The final outlier decision is computed by the logical AND of the outlier flags from both algorithms: $\textit{Final Outlier Flag} = \textit{IF Flag} \cap \textit{LOF Flag}$, where IF Flag is the result of applying Isolation Forest to the data,
LOF Flag is the result of applying Local Outlier Factor to the data. This ensemble approach ensures that the outlier is detected by both methods before it is flagged as a true outlier, thereby reducing false positives and increasing the reliability of the outlier detection method.

\section{Clustering Results}
    
Figure \ref{outlier}, shows examples of images from the MixedWM38 Validation set that were clustered using the RCAE feature extractor and DeepDPM non-parametric clustering approach. Outliers identified by the ensemble outlier-detection algorithm are highlighted with red outline. For outlier detection, we set the contamination prior parameters (\textit{hi\_cont}) of 0.20, robust cut scaling factor \textit{(k)} of 3 for both IF and LOF algorithms. For the adaptive \textit{k} parameter in LOF, we use a \textit{min\_k} and \textit{max\_k} of 10 and 50, respectively.

\bibliography{aaai2026}

\end{document}